\documentclass[10pt,twocolumn,letterpaper]{article}

\usepackage{iccv}              

\usepackage{booktabs, multirow} 
\usepackage{soul}
\usepackage{xcolor,colortbl} 
\usepackage{changepage,threeparttable} 

\usepackage{graphicx}    

\usepackage{amsmath}
\usepackage{tikz}
\usepackage{siunitx}

\usepackage{bm}
\definecolor{lightorange}{RGB}{254, 225, 200}
\definecolor{lightblue}{RGB}{181,199,229}

\definecolor{iccvblue}{rgb}{0.21,0.49,0.74}
\usepackage[pagebackref,breaklinks,colorlinks,allcolors=iccvblue]{hyperref}

\def\paperID{7269} 
\def\confName{ICCV}
\def\confYear{2025}

\title{GroundFlow: A Plug-in Module for Temporal Reasoning on 3D Point Cloud Sequential Grounding}

\author{
Zijun Lin$^{1,2}$ \quad
Shuting He$^{3}$ \quad
Cheston Tan$^{2}$ \quad
Bihan Wen$^{1}$ \\
\\
$^{1}$Nanyang Technological University \quad
$^{2}$Centre for Frontier AI Research, A*STAR \\
$^{3}$MoE Key Laboratory of Interdisciplinary Research of Computation and Economics, \\
Shanghai University of Finance and Economics
}

\begin{document}
\maketitle
\begin{abstract}
Sequential grounding in 3D point clouds (SG3D) refers to locating sequences of objects by following text instructions for a daily activity with detailed steps. Current 3D visual grounding (3DVG) methods treat text instructions with multiple steps as a whole, without extracting useful temporal information from each step. However, the instructions in SG3D often contain pronouns such as ``it'', ``here'' and ``the same'' to make language expressions concise. This requires grounding methods to understand the context and retrieve relevant information from previous steps to correctly locate object sequences. Due to the lack of an effective module for collecting related historical information, state-of-the-art 3DVG methods face significant challenges in adapting to the SG3D task. To fill this gap, we propose GroundFlow — a plug-in module for temporal reasoning on 3D point cloud sequential grounding. Firstly, we demonstrate that integrating GroundFlow improves the task accuracy of 3DVG baseline methods by a large margin (+7.5\% and +10.2\%) in the SG3D benchmark, even outperforming a 3D large language model pre-trained on various datasets. Furthermore, we selectively extract both short-term and long-term step information based on its relevance to the current instruction, enabling GroundFlow to take a comprehensive view of historical information and maintain its temporal understanding advantage as step counts increase. Overall, our work introduces temporal reasoning capabilities to existing 3DVG models and achieves state-of-the-art performance in the SG3D benchmark across five datasets. Codes are available at \href{https://github.com/Jimntu/GroundFlow}{\textcolor{iccvblue}{https://github.com/Jimntu/GroundFlow}}.
\end{abstract}    
\section{Introduction}
\label{sec:intro}

\begin{figure}[t]
    \centering
    \includegraphics[width=0.9\columnwidth]{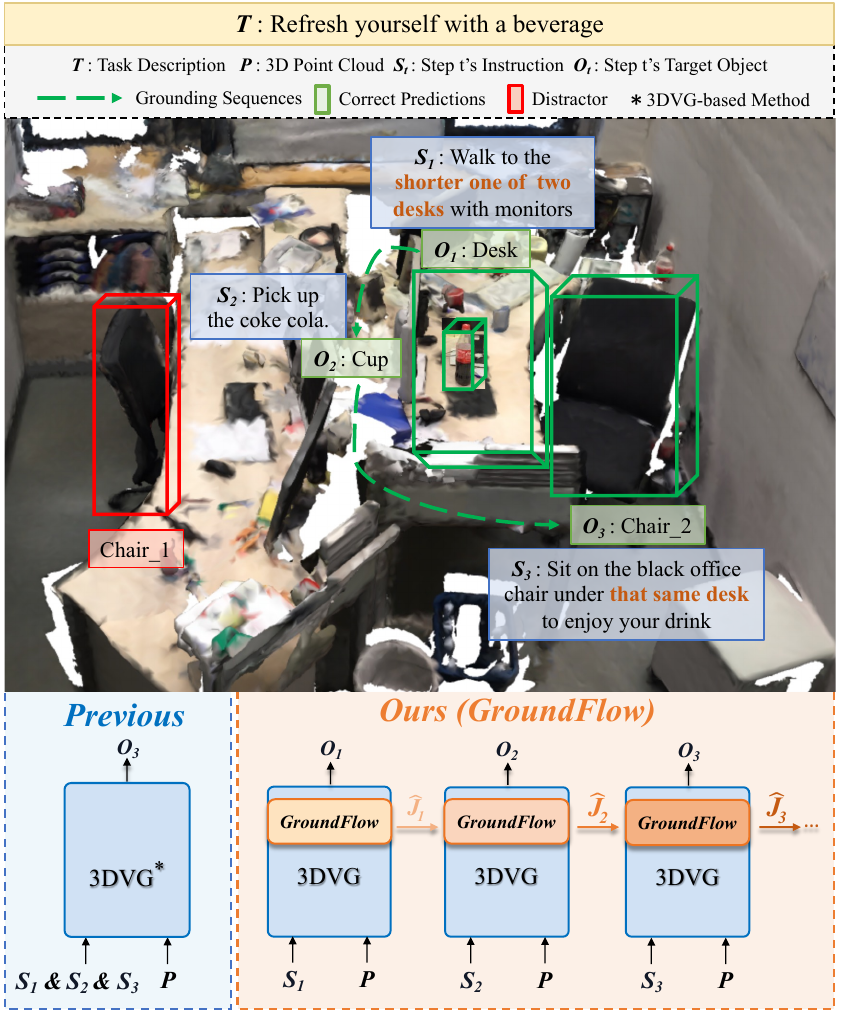}
      \caption{An example of SG3D task (above) and a comparison between previous visual grounding framework (bottom left) and our recurrent framework (bottom right) integrated with \textbf{GroundFlow} plug-in module to enhance temporal reasoning when localizing the target object in $S_3$. As shown, GroundFlow module's output $\hat{J}_t$ will be treated as input in the next step $t+1$.}
  \label{fig:recurrent}
\end{figure}

To understand and perform human instructions in the real world, robots should precisely identify the referred objects to complete complex tasks through natural language. 3D Visual Grounding (3DVG) \cite{instancerefer,zhao2021_3DVG_Transformer,feng2024naturally,yuan2024visual,wang2024g3lq} 
is a widely studied task that requires the agent to locate the target objects in 3D scenes based on a text instruction describing the object attributes in detail such as color, shape and spatial relationships. Recently, to advance grounding tasks toward more realistic applications, a new task called Sequential Grounding in 3D point clouds (SG3D) \cite{sequentialgrounding3d} has been introduced. Although this task shares similarities with 3DVG in grounding language information into the visual field, SG3D distinguishes itself by two characteristics --- \textit{sequential grounding} and \textit{task-driven}. As illustrated in the example in Figure \ref{fig:recurrent}, unlike 3DVG task, which provides detailed object information and grounds the target in a single step, SG3D requires agents to locate a sequence of objects in the correct order over multiple steps based on task-oriented instructions. This mirrors how humans instruct robots to carry out daily activities using step-by-step instructions, making the SG3D task more applicable to real-life scenarios. Therefore, properly addressing the SG3D problem is crucial for realizing practical embodied AI applications.

Current visual grounding models \cite{SegPoint, shi2024viewpoint, hsu2023ns3d,jin2023context,chen2023unit3d} show promising performance on many 3DVG datasets, such as ScanRefer \cite{scannet,scanrefer}, Sr3D, Nr3D \cite{referit_3d}, and Multi3DRefer \cite{multi3drefer}, but have poor generalization on the sequential grounding task \cite{sequentialgrounding3d}. The main reason for the huge performance gap between the two tasks is that current 3DVG methods are not designed to reason over historical information. As shown in Figure \ref{fig:recurrent}, 3DVG methods typically process all text instructions as a single, undifferentiated input, which works for traditional visual grounding tasks but is not suitable for the task-driven nature of SG3D. In SG3D, instructions rely heavily on contextual pronouns such as ``it'', ``here'', ``the same'', which are rarely seen in 3DVG tasks. These pronouns highlight the importance of effectively retrieving relevant information from past steps as the instructions progress. Simply concatenating all previous instructions to localize current step's target makes it difficult for the model to differentiate important history information from irrelevant details. Additionally, there is a trend of incorporating large language models into the 3D field due to their excellent reasoning capabilities and vast pre-trained knowledge \cite{leo,3dllm,zheng2024towards}. While 3D LLMs achieve state-of-the-art results in various 3D tasks, they still face significant difficulty adapting to the complex SG3D problem \cite{sequentialgrounding3d}. 

To address this, we design a recurrent framework, depicted in Figure \ref{fig:recurrent}. This framework sequentially takes each step instruction  and processes only the current step instruction as input rather than handling all prior text instructions simultaneously. In addition, we propose GroundFlow module, which can be built on top of the existing 3DVG methods to perform temporal fusion with previous step embeddings, improving the comprehension of history instructions. 

Furthermore, in the SG3D tasks, some step instructions depend on information from multiple previous steps. Take the task in Figure \ref{fig:recurrent} as an example, there are multiple chairs in the scene and to accurately identify the ``chair under that same desk'' for the last step, the model needs to retrieve relevant information about the ``shorter one of two desks'' mentioned in the first step. Attending only to the immediate previous step is insufficient and the problem would be more pronounced as the number of steps increase in the task, requiring the model to gather information from a longer timeline. Thus, carefully extracting useful information from both the short-term and long-term context is essential.

Humans tend to retrieve past information by first accessing recent activity and then integrating earlier details based on their relevance to the current context \cite{Neuroscience}. Accordingly, in the design of the GroundFlow module, we implement a memory structure mimicking how humans recall past information, selectively fusing short-term and long-term embeddings based on their correlation to the current embeddings. This approach allows the model to maintain perspective across time, efficiently leveraging both immediate and distant context as needed. 


In summary, we make the following contributions:

 \begin{itemize}
     \item We propose the GroundFlow module with a recurrent framework, which can be integrated into previous 3DVG baselines and introduce important temporal reasoning capabilities to them. Evaluated on SG3D benchmark, this approach results in significant improvements of up to 6.3\% and 10.2\% in task accuracy, as well as 3.8\% and 7.5\% in step accuracy for dual-stream and query-based models, which are two classic 3DVG methods. 
    \item We draw inspiration from how human retrieves the past information and effectively extract relevant short-term and long-term instructions based on current needs. This allows GroundFlow to improve the 3DVG baseline methods consistently as task steps increases.
    \item We achieve state-of-the-art performance on SG3D benchmark across five datasets without using large language model, outperforming 3D LLM by 2\% in both step accuracy and task accuracy.
 \end{itemize}
\section{Related Work}
\label{sec:formatting}
\subsection{3D Visual Grounding}

\begin{figure*}[h]
    \centering
    \includegraphics[width=\textwidth]{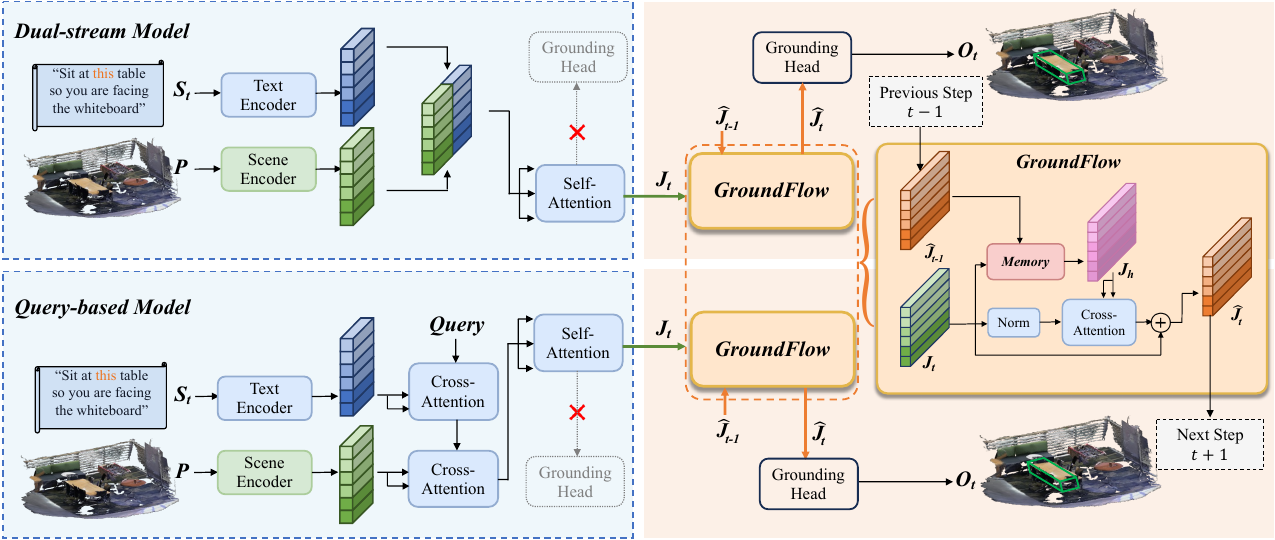}
      \caption{The overview of two 3DVG baseline models (\textcolor{lightblue}{blue} background) integrated with our proposed plug-in temporal fusion module --- \textbf{GroundFlow} (\textcolor{orange}{orange} background). Unlike baseline methods that directly use $J_t$ for grounding, GroundFlow integrates previous step information to produce $\hat{J}_t$, enhancing context-awareness in a recurrent framework.}
  \label{fig:model}
\end{figure*}

3D visual grounding (3DVG) aims to build connections between natural language and the 3D physical world \cite{yang2023exploiting,shi2024viewpoint,xu2024multi,feng2024naturally,yuan2024visual,wang2024g3lq,wang2023distilling}. Specifically, the 3DVG task requires agents to localize related target object(s) based on a single text instruction. In this emerging domain, several methods have been proposed to enhance agent's grounding capabilities in different aspects. For example, Zhang et al. \cite{multi3drefer} use M3DRef-CLIP, a CLIP-based \cite{radford2021learning} approach with contrastive learning, to ground text instructions to multiple 3D objects. Cai et al.  \cite{cai20223djcg} introduce a unified framework that jointly improves performance for dense captioning and visual grounding. Additionally, Guo et al. \cite{viewrefer} leverage the large language model to enrich input texts and adopt inter-view attention to address view discrepancy issues in 3DVG. Although these methods focus on different aspects, they can be broadly categorized into dual-stream models \cite{cai20223djcg, referit_3d,viewrefer,xu2024multi,chen_language_conditioned}, which process text and scene information separately before fusing them for visual grounding, and query-based models \cite{pq3d,butd,yang2023exploiting}, which initialize a query that sequentially incorporates text and scene information before the grounding stage. However, these methods are primarily designed for object-centric 3DVG tasks and face significant challenges when applied to sequential grounding tasks. In this work, we introduce a plug-in module, GroundFlow, which could be easily implemented on top of both dual-stream and query-based models, allowing existing 3DVG methods to transition smoothly to sequential grounding tasks.



\subsection{Temporal Context Learning}
Understanding and reasoning about the temporal information allow intelligent agents to retrieve useful past information and make logical predictions about current or future scenarios. This ability is thus widely explored in many research fields \cite{meng2021adafuse, zhou2018temporal, liang2024neural, wang2021sodtgnn}. Recurrent Neural Networks (RNNs) \cite{goodfellow2016deep} are commonly employed for temporal learning as they are designed to handle sequential data by maintaining a hidden state that captures dependencies over time. Chen et al. \cite{chen2021hamt} adopt History Aware Multimodal Transformer (HAMT) for the task of Vision-and-language navigation (VLN) to learn the temporal dynamics across panoramas of the navigation history. In video retrieval task, Shao et al. \cite{Shao_2021_WACV} propose temporal context aggregation to incorporate temporal information between frame-level features to better capture long-range dependency. Furthermore, in spatial-temporal video grounding, Gu et al. \cite{gu2024context} calculate the confidence score for each video frame in temporal refinement to filter out irrelevant instance context by a confidence threshold. In our work, we focus on referencing relevant short-term and long-term contextual information based on current-step instructions and introduce a recurrent framework that dynamically adapts to the sequential grounding and task-driven nature of the complex SG3D task.

\section{GroundFlow}

\newcommand{\circleM}{%
    \tikz[baseline=(M.base)] \node[draw, circle, inner sep=0.5pt] (M) {\textbf{\textit{S}}};%
}
In the SG3D task, we are provided with a 3D point cloud of an indoor scene \( \mathcal{P} \in \mathbb{R}^{N \times (3+C)}\) with ground-truth object masks, where $N$ denotes the number of points and $C$ represents the feature channels per point. The step-by-step instructions \( \mathcal{S} = \{ s_1, \ldots, s_n \} \) is given as textual input, where $n$ denotes the number of steps. The model is required to predict the object sequences \( \mathcal{O} = \{ o_1, \ldots, o_n \} \) based on the 3D point cloud and step instructions, meaning that the model needs to learn the mapping \( f : (\mathcal{P}, \mathcal{S}) \rightarrow \mathcal{O} \).

The challenges of the SG3D task, compared to other visual grounding tasks in 3D scenes, lie in grounding a sequence of targets given the implicit task-driven step instructions. Effectively retrieving relevant previous information is essential for solving this problem, yet current visual grounding methods lack a design for temporal fusion. In this section, we describe the technical details of our proposed plug-in module --- GroundFlow. Different from the general reasoning capabilities of large language models, GroundFlow specifically aims to equip current 3DVG models with temporal reasoning abilities. 


\subsection{Recurrent Framework with Temporal Fusion}
\label{subsec:recurrent}
Existing approaches in 3D visual grounding treat the entire text as a single input, which works well for most 3DVG datasets. However, these methods struggle to generalize in sequential grounding tasks, where the instructions are implicit and task-oriented, requiring the model to extract relevant information from previous steps conditioned on the current step instruction. Additionally, the average text length for a task in SG3D is 70.5 words, significantly longer than the 10 to 15 words typical in various 3DVG datasets. Simply concatenating all previous step instructions as input would introduce a substantial amount of redundant or irrelevant information, potentially degrading the model’s performance. Therefore, a recurrent framework with temporal fusion is proposed in GroundFlow to better capture the information at each step.

The proposed framework shown in Figure \ref{fig:model} is implemented on top of two types of 3DVG models --- dual-stream and query-based. Technically, the dual-stream models \cite{cai20223djcg, referit_3d,viewrefer,xu2024multi,chang2024mikasa} process the scene and text information separately and fuse them using a self-attention module, whereas the query-based models \cite{pq3d,butd,yang2023exploiting,chen_language_conditioned} sequentially process the scene and text embeddings to perform cross-attention with an initialized learnable query. Although the mechanisms for processing the input differ between the two models, both of them generate the joint embeddings $J_t$, which are intended to collectively represent the scene and text information in current step $t$. 

As shown in Figure \ref{fig:model}, our proposed recurrent framework takes the current step's joint embedding \(J_t\) and the previous step's GroundFlow module output \(\hat{J}_{t-1}\) as input, where the joint embedding with the hat  `$\hat{\phantom{x}}$'  denotes the output after performing temporal fusion in GroundFlow. Consequently, the module's output at the current step \(\hat{J}_t\) serves as input for the next step's inference. Since the joint embedding $J_t$ right before the grounding head is the only information required from 3DVG baselines to generate the corresponding $\hat{J}_t$ with important temporal information, our proposed recurrent framework is expected to adapt to a broader range of 3DVG model categories.

Equation \ref{eq1} shows how the module output $\hat{J}_t$ grasps the prior relevant information from history trajectory. Cross-attention is used to inject critical history information $J_h$ into the current step embeddings $J_t$. Notably, $J_h$ comprehensively integrates both short-term and long-term information through a novel memory structure, which will be further discussed in section \ref{subsec:memory}.

\begin{equation}
    \begin{cases}
   \hat{J}_{t} = J_t + \text{softmax} \left( \frac{J_t J_h^T}{\sqrt{H}} \right) J_h , & t \in [2, n] \\
    \hat{J}_{t} = J_t, & t = 1
    \end{cases},
\label{eq1}  
\end{equation}
where \( J_t \in \mathbb{R}^{T \times H} \) contains $T$ tokens and $H$ dimension for each token. After the cross-attention, $\hat{J}_{t}$ captures both the current step's scene-text relationship and the information from the relevant previous steps. For the first step, since there is no historical information, $\hat{J}_{t}$ directly takes the value of $J_t$'s similar to traditional 3DVG methods. 

Afterwards, $\hat{J}_{t}$ is fed into the grounding head to predict the target object $O_t$ for current step $t$. The structure of the grounding head in 3DVG models typically consists of several MLP layers. Hence, our recurrent framework introduces the temporal reasoning capabilities for previous visual grounding models without hindering their predictions on the 3DVG task.



\subsection{Memory Structure}
\label{subsec:memory}


Directly applying the immediate previous instruction $\hat{J}_{t-1}$ to fuse temporally with the current embedding $J_t$ only allows the output $\hat{J}_t$ to contain relevant information from the immediate last step without the consideration of long-term information. Earlier information tends to fade as the step evolves, but there are cases where important information is mentioned in initial step instructions. How to effectively retrieve instructions for both short-term and long-term memory conditioned on current needs is challenging but crucial for GroundFlow to achieve more robust temporal reasoning. 

\begin{figure}[t]
    \centering
    \includegraphics[width=\columnwidth]{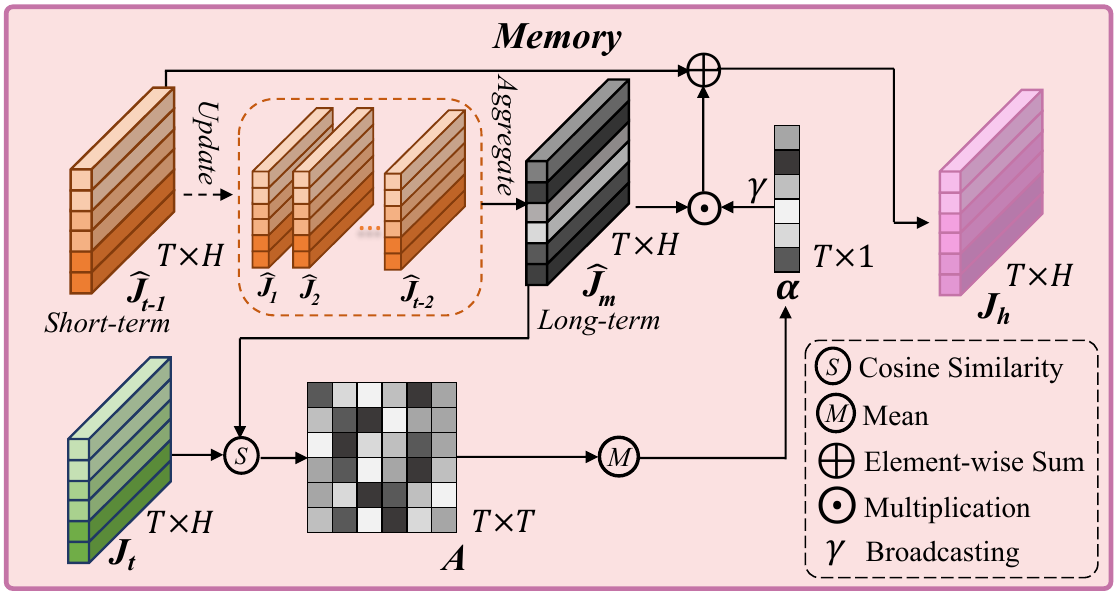}
      \caption{ Detailed illustration of Memory component in GroundFlow, which enables the module to extract relevant information of both \textbf{short-term} ($\hat{J}_{t-1}$) and \textbf{long-term} ($\hat{J}_m$) effectively. The shape of each tensor is labeled adjacent to it. The dotted arrow represents the update to the long-term memory for the next step, which will be performed \textbf{after} the current step $t$ is completed.}
  \label{fig:memory}
\end{figure}

To address this, we further implement a memory structure in GroundFlow. A detailed illustration of the memory component is shown in Figure \ref{fig:memory}. The memory takes the current step's joint embedding $J_t$ and the last step's GroundFlow output $\hat{J}_{t-1}$ as input. All previous step information is processed as long-term information within the memory structure to generate the history embedding $J_h$:
\begin{equation}
    \begin{cases}
   J_h = \text{Memory}(\hat{J}_{t-1}, J_t), & t \in [3, n] \\
    J_h = \hat{J}_{t-1}, & t = 2
    \end{cases}.
\label{eq2}  
\end{equation}
It is worth noting that the memory component only starts functioning after the second step, as there is no long-term instruction prior to it. For each task step, tokens are padded to the maximum length, and a padding mask is created to help the model ignore padding tokens during processing.

In the memory structure, the recent GroundFlow output $\hat{J}_{t-1}$ is regarded as short-term memory, while $\hat{J}_m$ is considered long-term memory, which aggregates all the processed joint embeddings from the first step up to two steps prior:
\begin{equation}
\hat{J}_m =  \sum_{t=1}^{t-2} \hat{J}_{t},
\end{equation}
where \( \hat{J}_m \in \mathbb{R}^{T \times H} \) contains $T$ tokens, with each token having a dimension $H$, matching the shape of $\hat{J}_{t}$. As indicated by the dotted arrow in Figure \ref{fig:memory}, $\hat{J}_m$ is updated (i.e.,  $\hat{J}_m = \hat{J}_m + \hat{J}_{t-1}$) for next step after completing the current memory inference.

Given that not all the previous step's long-term information is relevant to the current prediction, naively using entire history embeddings indiscriminately is not an optimal solution. Therefore, we adopt cosine similarity to extract the most relevant information in $\hat{J}_m$ based on the current joint embedding $J_t$:
\begin{equation}
\mathbf{A}_{i, j} = \circleM \left( \hat{J}_m, J_t \right),
\end{equation}
where $\circleM$ denotes cosine similarity and \( \mathbf{A}_{i, j} \in \mathbb{R}^{T \times T} \). Each row of $\mathbf{A}_{i, j}$ indicates the importance of each token in $\hat{J}_m$ to the current joint embedding $J_t$. The value of a specific row is expected to be higher if the corresponding tokens in $\hat{J}_m$ contain relevant information for $J_t$.

After obtaining the importance score for each token in the long-term memory $\hat{J}_m$, we average all the scores to get a single value for each token for subsequent calculations. This gives each token in $\hat{J}_m$ a  relative importance score with respect to $J_t$:
\begin{equation}
\alpha = \frac{1}{T} \sum_{j=1}^{T} \mathbf{A}_{i, j},
\end{equation}
where \( \alpha \in \mathbb{R}^{T \times 1} \). This approach allows the model to treat each history token differently based on its relevance to the current information.

The ultimate history embedding $J_h$ combines both short-term and long-term information. We achieve this by summing the original short-term memory $\hat{J}_{t-1}$ with the weighted long-term memory $\hat{J}_m \odot \gamma(\alpha)$ together. The approach is inspired by the human memory retrieval process, which intuitively reflects on the most recent activity and the most relevant past experiences. The final equation for obtaining the history information $J_h$ is shown as follows: 
\begin{equation}
    J_h =\hat{J}_{t-1} + \hat{J}_m \odot \gamma(\alpha),
\label{eq:memory_sum}
\end{equation}
where $\odot$ is the Hadamard product (i.e., element-wise multiplication) and $\gamma$ is the broadcasting function. 

Subsequently, cross-attention is applied between the current joint embedding $J_t$ and the history information $J_h$, as shown in Equation \ref{eq1}, yielding $\hat{J}_{t}$ as the final representation before the grounding head. This representation incorporates the current step embedding with rich contextual understanding. Overall, the memory structure provides the model with a comprehensive view of the prior trajectory and extract relevant historical information, further enhancing the temporal reasoning capabilities of GroundFlow.

\subsection{Training Objective}
\label{subsec:training}
Following the SG3D benchmark \cite{sequentialgrounding3d}, we use the same cross-entropy loss to optimize the dual-stream model and the query-based model. As defined in Equation \ref{eq:loss}, the loss compares the predicted object score \(f(\mathcal{P}, \mathcal{S}) \) and the ground truth score $\mathcal{O}$. For the 3D LLM LEO \cite{leo}, which is state-of-the-art method in SG3D benchmark, we follow the original approach. In addition to the loss of token predictions when pre-trained on other datasets, an extra cross-entropy loss is incorporated to fine-tune the model on SG3D data.
\begin{equation}
\label{eq:loss}
\mathcal{L}_{\text{grd}} = \mathbb{E}_{(\mathcal{P}, \mathcal{S}, \mathcal{O}) \sim \mathcal{D}} \text{CrossEntropy} \left( f(\mathcal{P}, \mathcal{S}), \mathcal{O} \right).
\end{equation}

\section{Experiments}
\subsection{Dataset and Evaluation Metrics}
\textbf{Dataset.} GroundFlow is evaluated on the newly proposed benchmark --- SG3D \cite{sequentialgrounding3d}, which is specifically designed for the task of sequential grounding in 3D point clouds. The benchmark utilizes real-world scenes from the SceneVerse \cite{jia2024sceneverse}, incorporating indoor scans from 5 different datasets --- ScanNet \cite{scannet}, 3RScan \cite{Wald2019RIO}, MultiScan \cite{mao2022multiscan}, ARKitScenes \cite{dehghan2021arkitscenes} and HM3D \cite{ramakrishnan2021hm3d}. Detailed statistics of the SG3D benchmark are provided in Table \ref{tab:dataset}.


\begin{table*}[!t]\centering
\setlength{\arrayrulewidth}{0.8pt}
\scriptsize
\begin{tabular}{c|c|cc|cc|cc|cc|cc|cc}
\hline 
\multirow{2}{*}{Model Type} &\multirow{2}{*}{Method} &\multicolumn{2}{c|}{ScanNet} &\multicolumn{2}{c|}{3RScan} &\multicolumn{2}{c|}{MultiScan} &\multicolumn{2}{c|}{ARKitScenes} &\multicolumn{2}{c|}{HM3D} &\multicolumn{2}{c}{Overall} \\
& &s-acc & t-acc & s-acc & t-acc & s-acc & t-acc & s-acc & t-acc & s-acc & t-acc & s-acc & t-acc \\
\hline
\multirow{2}{*}{LLM-based}&
GPT4 + PointNet++ (Zero-shot) \cite{sequentialgrounding3d}&42.6&10.9&25.5&2.4&27.0&0.0&27.6&6.0&20.8&7.7&27.3&7.6\\
& LEO (3DLLM) \cite{leo} & 61.2 & 25.7 & 55.8 & 16.0 & 52.7 & 7.6 & 69.6 & 41.5 & 61.5 & 35.7 & 62.8 & 34.1 \\
\hline
\multirow{4}{*}{Dual-stream} & 3D-VisTA \cite{3dvista} & 60.1 & 24.7 & 52.7 & 13.5 & 47.6 & 7.0 & 68.4 & 37.8 & 57.5 & 30.6 & 60.3 & 28.8 \\
& \cellcolor{lightorange}3D-VisTA\textbf{+ GroundFlow} & \cellcolor{lightorange}63.0 & \cellcolor{lightorange}26.6 & \cellcolor{lightorange}56.8 & \cellcolor{lightorange}21.7 & \cellcolor{lightorange}57.1 &\cellcolor{lightorange} 14.0 & \cellcolor{lightorange}71.9 & \cellcolor{lightorange}46.0 & \cellcolor{lightorange}62.3 & \cellcolor{lightorange}36.9 & \cellcolor{lightorange}64.1 & \cellcolor{lightorange}35.1 \\
& MiKASA \cite{chang2024mikasa} & 57.8 & 19.4 & 53.0 & 10.9 & 48.7 & 2.3 & 67.1 & 35.7 & 57.3 & 30.1 & 60.8 & 31.9 \\
& \cellcolor{lightorange} MiKASA \textbf{+ GroundFlow} & \cellcolor{lightorange}62.7&\cellcolor{lightorange}\textbf{28.9}&\cellcolor{lightorange}58.9&\cellcolor{lightorange}17.4&\cellcolor{lightorange}54.0&\cellcolor{lightorange}11.6&\cellcolor{lightorange}70.2&\cellcolor{lightorange}42.9&\cellcolor{lightorange}61.8&\cellcolor{lightorange}36.2&\cellcolor{lightorange}63.5&\cellcolor{lightorange}34.2 \\
\hline
\multirow{4}{*}{Query-based} & PQ3D \cite{pq3d} & 53.7 & 17.9 & 50.2 & 9.9 & 43.5 & 4.7 & 64.9 & 32.0 & 56.9 & 30.6 & 57.3 & 25.9 \\
& \cellcolor{lightorange} PQ3D \textbf{+ GroundFlow} & \cellcolor{lightorange}62.0 & \cellcolor{lightorange}28.2 & \cellcolor{lightorange}\textbf{60.1} & \cellcolor{lightorange}21.0 & \cellcolor{lightorange}51.3 & \cellcolor{lightorange}7.0 & \cellcolor{lightorange}\textbf{73.0} & \cellcolor{lightorange}\textbf{48.1} &\cellcolor{lightorange} \textbf{63.6} & \cellcolor{lightorange}\textbf{38.0} & \cellcolor{lightorange}\textbf{64.8} & \cellcolor{lightorange}\textbf{36.1} \\
& Vil3DRel \cite{chen_language_conditioned} & 59.3&19.9&55.9&15.2&50.9&4.7&69.3&38.6&58.7&31.0&61.1&28.6\\
& \cellcolor{lightorange} Vil3DRel \textbf{+ GroundFlow} &\cellcolor{lightorange}\textbf{63.1}&\cellcolor{lightorange}27.8&\cellcolor{lightorange}58.8&\cellcolor{lightorange}\textbf{22.5}&\cellcolor{lightorange}\textbf{57.6}&\cellcolor{lightorange}\textbf{20.9}&\cellcolor{lightorange}72.4&\cellcolor{lightorange}45.1&\cellcolor{lightorange}62.3&\cellcolor{lightorange}36.6&\cellcolor{lightorange}64.4&\cellcolor{lightorange}35.2 \\
\hline
\end{tabular}
\caption{Comparisons on SG3D benchmark across five datasets. The values for the metrics s-acc (step accuracy) and t-acc (task accuracy) are expressed as percentages (\%). The methods with our proposed modules are highlighted in \textcolor{orange}{orange} and the numbers in bold represent the best performance in given dataset and metrics.}
\label{tab:SG3D}
\end{table*}

\noindent \textbf{Evaluation Metrics.} As defined in SG3D benchmark \cite{sequentialgrounding3d}, all models' grounding performances is evaluated based on two key metrics: step accuracy (s-acc) and task accuracy (t-acc). Step accuracy is calculated as the average accuracy of the model in predicting the target object across all individual steps $S$, while task accuracy refers to the grounding accuracy over the total number of tasks $T$. For task accuracy, a sample is considered correct if the predicted sequence of objects for each step matches the ground-truth sequence.


\subsection{Implementation Details}
\label{subsec:hyper}
For a fair comparison, all methods follow the same hyperparameter settings. The models are trained for 50 epochs with batch size of 32 and evaluated on the last epoch using evaluation split of the SG3D benchmark. AdamW optimizer is employed with the learning rate of 1e-4 and weight decay of 0.05 for optimization, while $\beta_{1}$, $\beta_{2}$ are set to 0.9 and 0.999, respectively. Due to GPU memory constraints, the batch size for LEO is reduced to 16. All other training details for the baselines strictly follow the original paper’s settings. All 3DVG methods integrated with GroundFlow (only 22M \#params) are able to deploy on single NVIDIA 24GB A5000 GPU. Kindly refer to Table \ref{tab:compute} for comparisons of computational efficiency.

\subsection{Comparison on SG3D Benchmark}

\noindent \textbf{Comparison with 3DVG methods.} We integrate GroundFlow into two classic 3DVG models -- dual-stream and query-based. Multiple state-of-the-art 3DVG models (3D-VisTA \cite{3dvista} and MiKASA \cite{chang2024mikasa} for dual-stream, PQ3D \cite{pq3d} and Vil3DRel \cite{chen_language_conditioned} for query-based) are selected to represent these two categories. To demonstrate the effectiveness of GroundFlow, the experimental results of these approaches before and after introducing GroundFlow will be compared. It is important to note that, beyond these 3DVG approaches, GroundFlow can serve as a plug-in module for a broader range of visual grounding methods, as it only requires the joint embedding right before the grounding head as input to perform temporal fusion within a recurrent framework.

Table \ref{tab:SG3D} highlights that the previous 3DVG methods struggle to generalize to the SG3D benchmark. Their degraded performance is particularly reflected in their overall task accuracy, with three of the models are falling below 30\%. These results are expected due to the absence of a temporal reasoning module. On the other hand, significant performance improvements can be observed when these models are integrated with GroundFlow, as shown in the rows highlighted in orange. Both the step accuracy and the task accuracy see substantial boosts in SG3D benchmark for all the 3DVG methods. Specifically, the overall performance in the SG3D benchmark of 3D-VisTA increases by 3.8\% in step accuracy and 6.3\% in task accuracy. For PQ3D, the improvements are even more pronounced, with gains of 7.5\% and 10.2\%, respectively and the task accuracy improves by 1.2 to 2.1 times across five datasets with the introduction of GroundFlow. These remarkable improvements validate our findings that effective temporal fusion is crucial for addressing the sequential grounding challenge. 

\noindent \textbf{Comparison with LLM-based methods.} Following \cite{sequentialgrounding3d}, an LLM with the 3D object classifier PointNet++ \cite{qi2017pointnetplusplus} is tested in a zero-shot manner. GPT-4 receives scene information, including each object's ID, size, position, and the step instruction, while PointNet++, pre-trained on ScanNet, is used to predict the semantic labels for each object. The results in the first row of the table demonstrate poor zero-shot performance. Furthermore, the state-of-the-art 3D large language model, LEO, after fine-tuning on the SG3D benchmark is also compared. It is pre-trained on an extensive range of 3D tasks, including object captioning \cite{li2023cap3d, yang2021object}, object referring \cite{instancerefer, feng2022reftransformer, he2021transrefer3d}, 3D QA \cite{gordon2018iqa, das2018embodiedqa,yu2022multi} and 3D navigation \cite{krantz2020beyond, ilker2022navigation}, achieving top performance across various 3D point cloud benchmarks. In fine-tuning stage, LEO predicts a special $[GRD]_t$ token at each step $t$, which is concatenated with object tokens and passed to the grounding head to predict the target object $O_t$. It is shown that LEO has decent results in SG3D benchmark and performs better than previous 3DVG methods due to its Internet-scale pre-trained knowledge. 


However, the 3DVG methods combined with our proposed GroundFlow module outperform LEO across all five datasets, setting new state-of-the-art performance on SG3D benchmark. With the integration of GroundFlow, the step accuracy and task accuracy of the dual-stream model 3D-VisTA are 1.3\% and 1\% higher than LEO, respectively, while the query-based PQ3D surpasses LEO by 2\% for both metrics. Therefore, LLM-based solutions excel at common-sense reasoning but may struggle with temporal knowledge, while GroundFlow focuses on modeling temporal dependencies in sequential grounding and selectively integrates historical information, leading to the better performance compared to 3D LLM. 

Overall, GroundFlow, as a plug-in module, successfully enhances the capability of 3DVG models to handle temporal information, enabling them to seamlessly transition to sequential grounding tasks.

\subsection{Ablation Study}

\noindent \textbf{Comparisons of different temporal fusion methods.} To evaluate the effectiveness of our proposed temporal fusion module, GroundFlow, we compare its results on 3D-VisTA and PQ3D with other established approaches such as LSTM \cite{LSTM}, GRU \cite{GRU} and Transformer \cite{vaswani2017attention}. 

\begin{figure*}[h]
    \centering
    \includegraphics[width=0.95\textwidth]{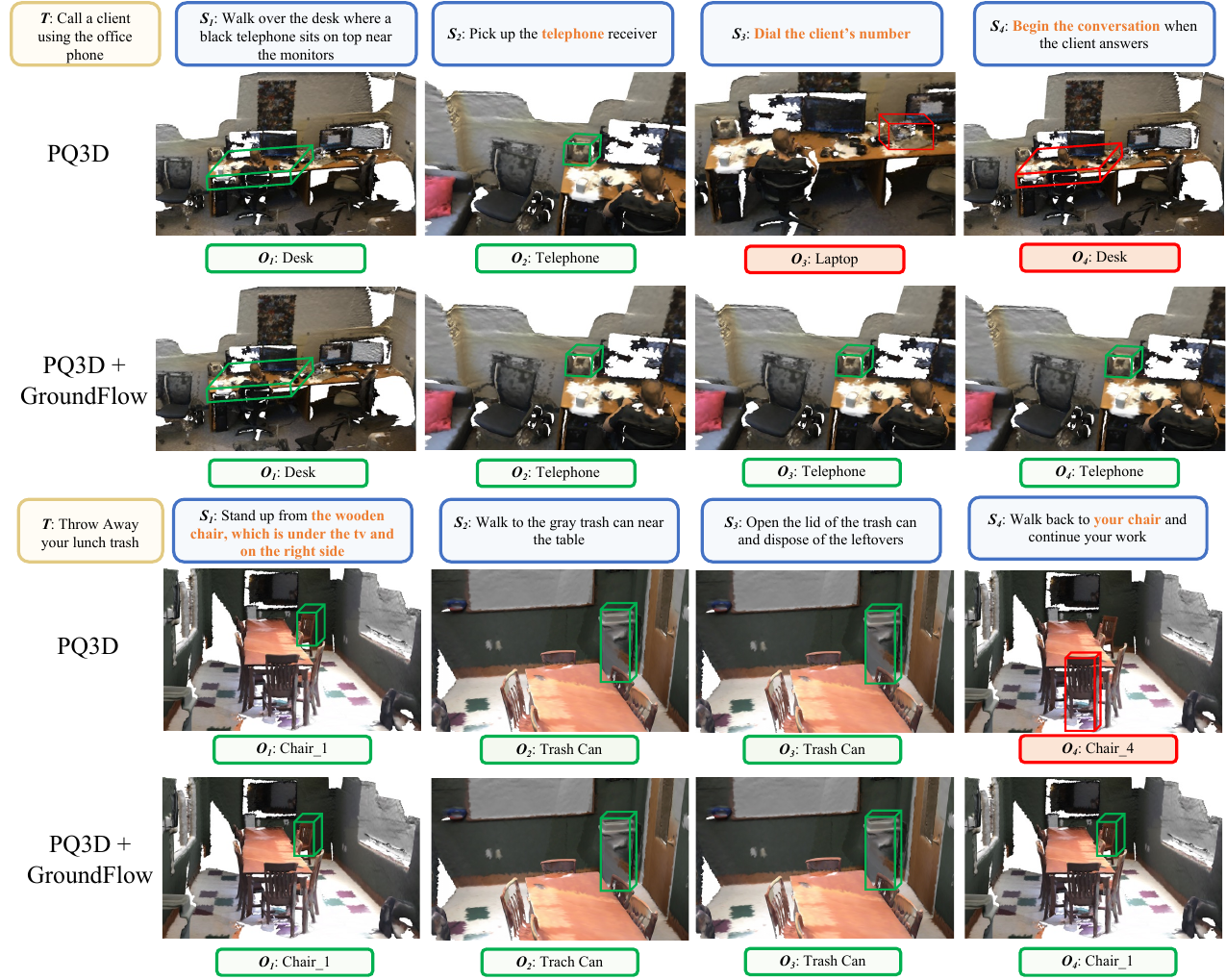}
      \caption{Visualization results from PQ3D and PQ3D\texttt{+}GroundFlow. $T$ represents the task description, $S_t$ and $O_t$ denote the step instruction and corresponding referred target object in step $t$. \textcolor{red}{Red} are wrong predictions and \textcolor{green}{green} are correct predictions.}
  \label{fig:visual}
\end{figure*}

As shown in Table \ref{tab:temporal_fusion_results}, GroundFlow demonstrates more effective improvements compared to classic temporal fusion methods. This advantage could stem from the limitations of existing methods: LSTM or GRU tends to forget long-term information. While transformers use attention mechanisms to capture long-range dependencies, they treat each step indiscriminately. As a result, they may still lose crucial historical information that is important at current step $t$ but not relevant at previous step $t-1$. Since previous step embeddings do not attend to this lost information, it cannot be carried forward to subsequent steps, even if it is essential for the current prediction. This issue is common in SG3D, where the current instruction often refers to objects from multiple past steps while intermediate instructions are irrelevant, as illustrated in the second example of Figure \ref{fig:visual}.

To address these limitations, the memory component in GroundFlow computes similarity scores to selectively retrieve and integrate context-specific past information based on its relevance to the current step, leading to superior performance enhancements, especially in task accuracy.


\begin{table}[t]
\centering
\setlength{\arrayrulewidth}{0.8pt}
\scriptsize
\begin{tabular}{c|c|c|c}
\hline
Models & Temporal Fusion Methods & s-acc \hspace{0.15cm} t-acc &\( \Delta \)s-acc \( \Delta \)t-acc \\
\hline
\multirow{4}{*}{3D-VisTA} & LSTM & 61.4 \hspace{0.15cm} 29.5 & +1.1 \hspace{0.15cm} +0.7 \\
 & GRU & 62.0 \hspace{0.15cm} 28.8 & +1.7 \hspace{0.15cm} +0.0\\
 & Transformer & 62.9 \hspace{0.15cm} 33.5 & +2.6 \hspace{0.15cm} +4.7 \\
 & \cellcolor{lightorange}\textbf{GroundFlow} & \cellcolor{lightorange}\textbf{64.1} \hspace{0.15cm} \cellcolor{lightorange}\textbf{35.1} & \cellcolor{lightorange}\textbf{+3.8} \hspace{0.15cm} \cellcolor{lightorange}\textbf{+6.3} \\
\hline
\multirow{4}{*}{PQ3D} & LSTM & 63.1 \hspace{0.15cm} 30.8 & +5.8 \hspace{0.15cm} +4.9 \\
 & GRU & 63.8 \hspace{0.15cm} 30.7 & +6.5 \hspace{0.15cm} +4.8 \\
 & Transformer & 63.4 \hspace{0.15cm} 33.6 & +6.1 \hspace{0.15cm} +7.7 \\
 & \cellcolor{lightorange}\textbf{GroundFlow} & \cellcolor{lightorange}\textbf{64.8} \hspace{0.15cm} \cellcolor{lightorange}\textbf{36.1} &
 \cellcolor{lightorange}\textbf{+7.5} \hspace{0.03cm} \cellcolor{lightorange}\textbf{+10.2} \\
\hline
\end{tabular}
\caption{Comparison of temporal fusion methods for 3D-VisTA and PQ3D. The $\Delta$ improvement in the last two columns is relative to the original 3DVG baselines.}
\label{tab:temporal_fusion_results}
\end{table}

\noindent\textbf{Improvements in t-acc over task steps.} In the SG3D benchmark, the step count of a task ranges from 2 to 10. To investigate whether GroundFlow can consistently improve task accuracy of the 3DVG methods in more challenging scenarios with a high number of steps, we create seven subsets from the evaluation split. These subsets contain tasks with step counts ranging from 2 to 7. Tasks with step counts greater than 7 are combined into one subset, as creating separate subsets for them would result in fewer than 100 tasks, which may not be sufficient to reliably reflect performance. 

\begin{figure}[t]
    \centering
    \includegraphics[width=0.8\columnwidth]{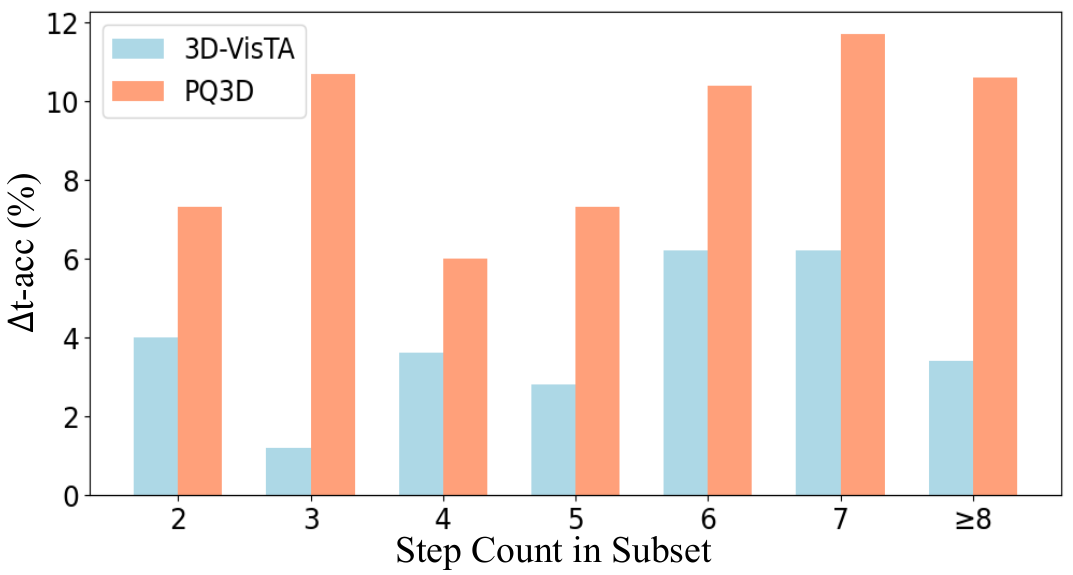}
      \caption{Improvements after GroundFlow module is integrated in terms of task accuracy of 3D-VisTA and PQ3D across different step count subsets.}
  \label{fig:step}
\end{figure}

As shown in Figure \ref{fig:step}, the introduction of GroundFlow improves the performance of all subsets with different step counts, with the query-based method PQ3D showing greater performance gains compared to the dual-stream method 3D-VisTA. Notably, task accuracy improvements for PQ3D exceed 10\% across all subsets with step counts greater than six, further highlighting GroundFlow's effectiveness in retrieving long-term information in the challenging high-step-count subsets.
\begin{table}[t]
\centering
\setlength{\arrayrulewidth}{0.8pt}
\scriptsize
\begin{tabular}{c|c|c|c|c}
\hline
Models & Short-term & Long-term & s-acc \hspace{0.07cm} t-acc &\( \Delta \)s-acc   \( \Delta \)t-acc \\
\hline
\multirow{5}{*}{3D-VisTA} & $\hat{J}_{t-1}$ & N.A. & 63.4 \hspace{0.15cm} 34.2 &+3.1 \hspace{0.15cm} +5.4  \\
 & N.A. & $\hat{J}_{t-1}+\hat{J}_m$ & 64.1 \hspace{0.15cm} 34.6 &+3.8 \hspace{0.15cm} +5.8 \\
 & $J_{t-1} $ & $\hat{J}_m$ & 64.0 \hspace{0.15cm} 34.7&+3.7 \hspace{0.15cm} +5.9 \\
 & $\hat{J}_{t-1}$ &  ${J}_m$  & 64.1 \hspace{0.15cm} 34.6&+3.8 \hspace{0.15cm} +5.8 \\
 & \cellcolor{lightorange} $\hat{J}_{t-1}$ 
 & \cellcolor{lightorange}$\hat{J}_m$ 
 & \cellcolor{lightorange}\textbf{64.1} 
 \hspace{0.15cm} \cellcolor{lightorange}\textbf{35.1} 
  & \cellcolor{lightorange}\textbf{+3.8}  \hspace{0.15cm}
  \cellcolor{lightorange}\textbf{+6.3}\\
\hline
\multirow{5}{*}{PQ3D} & $\hat{J}_{t-1}$ & N.A. & 64.3 \hspace{0.15cm} 35.7 &+7.0 \hspace{0.15cm} +9.8\\
 & N.A. & $\hat{J}_{t-1}+\hat{J}_m$ & 63.6 \hspace{0.15cm} 34.5&+6.3 \hspace{0.15cm} +8.6 \\
 & $J_{t-1} $ & $\hat{J}_m$ & 63.3 \hspace{0.15cm} 33.3 &+6.0 \hspace{0.15cm} +7.4\\
 & $\hat{J}_{t-1}$ &  ${J}_m$  & 63.4 \hspace{0.15cm} 34.6 &+6.1 \hspace{0.15cm} +8.7\\
 & \cellcolor{lightorange}$\hat{J}_{t-1}$ &  \cellcolor{lightorange}$\hat{J}_m$  & \cellcolor{lightorange}\textbf{64.8} \hspace{0.15cm} \cellcolor{lightorange}\textbf{36.1}   & \cellcolor{lightorange}\textbf{+7.5}  \hspace{0.05cm}
  \cellcolor{lightorange}\textbf{+10.2} \\
\hline
\end{tabular}
\caption{Comparison of different short-term and long-term memory settings for 3D-VisTA and PQ3D. Aligned with section \ref{subsec:memory}, $\hat{J}_m$ denotes the aggregation of the GroundFlow output $\hat{J}_t$ from the first step to two steps prior while $J_m$ denotes the summation of original joint embeddings $J_t$, (i.e., $\hat{J}_m =  \sum_{t=1}^{t-2} \hat{J}_{t},{J}_m =  \sum_{t=1}^{t-2} {J}_{t}$).}
\label{tab:memory}
\end{table}

\noindent \textbf{Impact of different memory configurations.}
In GroundFlow, the history embeddings $J_h$ accumulate rich information from both short-term and long-term sources by summing them up, as described in Equation \ref{eq:memory_sum}. To explore the impact of different memory configurations, we compare the various settings of short-term and long-term information within the memory component. In Table \ref{tab:memory}, the performance without one of the memory parts is presented in the first and second rows. Additionally, we assess whether embeddings before or after the application of GroundFlow yield better results, with their outcomes shown in the third and fourth rows. The final configuration adopted in GroundFlow, highlighted in orange, is shown in the fifth row for both models. 

Results in Table \ref{tab:memory} show that both short-term and long-term memory are crucial for effective memory retrieval, as the absence of either term results in a performance downgrade. Furthermore, the memory component using the joint embeddings after performing temporal fusion in GroundFlow yields better results, demonstrating that the temporal reasoning capabilities provided by GroundFlow enable the memory structure to capture important temporal information. Therefore, we use $\hat{J}_{t-1}$ as short-term memory and $\hat{J}_m$ as long-term memory in our final implementation, leading to the best performance.

\subsection{Qualitative Visualization}

Figure \ref{fig:visual} shows the visualization results of one of our baseline methods PQ3D, alongside its integration with GroundFlow. In the first example, to localize the correct target for the last two step instructions ``Dial the client's number'' and ``Begin the conversation'', models need to understand the context that the telephone has been picked up in previous steps. It is shown that PQ3D fails to correctly choose the target ``Telephone'', while PQ3D+GroundFlow makes the correct predictions of ``Telephone'' for both steps. Therefore, with the help of our proposed GroundFlow module, 3DVG methods can capture contextual information and exhibit essential temporal reasoning capabilities, generalizing to challenging task-oriented sequential grounding tasks. 

In the second example, there are multiple chairs in the scene and the last step instruction,  ``Walk back to your chair'', refers to the chair that is ``under the tv and on the right side'' mentioned in the first step. This requires the model to retrieve earlier information beyond just short-term memory. Unlike PQ3D, which incorrectly selects another chair, PQ3D integrated GroundFlow consistently identifies the correct chair referenced in the first step. These results highlight that the memory component in GroundFlow enables the model to retain important context over time, allowing it to accurately retrieve and apply information from both recent and earlier steps. Kindly refer to Figure \ref{fig:supp_visual} for more qualitative comparisons.

\section{Conclusion}
We propose GroundFlow, a plug-and-play module with a recurrent framework that introduces temporal reasoning capabilities to existing 3D visual grounding models, enabling them to seamlessly adapt to complex and challenging sequential grounding tasks. The memory structure within the GroundFlow module effectively retrieves relevant short-term and long-term historical information based on current needs. The extensive experiments show that, with the integration of GroundFlow, the performance of both classic types of 3DVG models --- dual-stream and query-based, improve significantly. Their results surpass the 3D large language model pre-trained in various 3D tasks, achieving consistent state-of-the-art performance on SG3D benchmark.

\section*{Acknowledgements}

Zijun Lin is supported by the Agency for Science, Technology and Research (A*STAR) Computing and Information Science (ACIS) Scholarship. 
Shuting He is sponsored by Shanghai Pujiang Programme 24PJD030 and Natural Science Foundation of Shanghai 25ZR1402138.
This research is supported in part by A*STAR SERC CRF funding to C.T., and in part by A*STAR IAF-ICP Programme I2501E0041 and the Schaeffler-NTU Corporate Lab (SHARE@NTU). The work was done at Rapid-Rich Object Search (ROSE) Lab, School of Electrical \& Electronic Engineering, Nanyang Technological University.

{   
    \small
    \bibliographystyle{ieeenat_fullname}
    \bibliography{main}
}

\clearpage
\setcounter{page}{1}
\maketitlesupplementary
\setcounter{section}{0} 
\renewcommand{\thesection}{\Alph{section}}

\section{SG3D Benchmark Statistics}

In Table \ref{tab:dataset}, we show the detailed dataset statistics in the SG3D benchmark. There are 22,346 tasks and 112,336 steps in total, with an average of 5.03 steps per task, an average length of 12.7 words per step instruction, and 70.5 words per task.  All five datasets in SG3D were split into training and evaluation sets. 
\begin{table}[h]
\centering
\setlength{\arrayrulewidth}{0.8pt}
\scriptsize
\begin{tabular}{ccccc}
\hline
\textbf{Dataset} & \#\textbf{$P$} & \#\textbf{$O$/$P$} & \#\textbf{$T$} & \#\textbf{$S$} \\
\hline
ScanNet     & 693   & 30.7 & 3,174  & 15,742  \\
3RScan       & 472   & 31.5 & 2,194  & 11,318   \\
MultiScan     & 117   & 40.8 & 547    & 2,683   \\
ARKitScenes   & 1,575 & 12.1 & 7,395  & 39,887  \\
HM3D      & 2,038 & 31.0 & 9,036  & 42,706  \\
\hline
\textbf{Total}    & 4,895 & 25.1 & 22,346 & 112,336 \\
\hline
\end{tabular}
\caption{Dataset statistics of SG3D benchmark. \#$P$, \#$O$/$P$, \#$T$ and \#$S$ represent the number of 3D point cloud scenes, the average number of candidate objects per scene, the number of tasks and the number of steps, respectively.}
\label{tab:dataset}
\end{table}

\section{Model Computational Complexity}
In Table \ref{tab:compute}, 3D-VisTA and PQ3D are selected to integrate with the GroundFlow module as two examples to demonstrate a better trade-off between accuracy and speed. With only a marginal increase in inference time (approximately 0.4ms for 3D-VisTA and 0.1ms for PQ3D), both step accuracy and task accuracy improve significantly. Additionally,  since GroundFlow has only 22M parameters, all 3DVG + GroundFlow experiments can be efficiently deployed on a single NVIDIA 24GB A5000 GPU.

\begin{table}[h]
\centering
\scriptsize
\begin{tabular}{c|c|c|cc}
\hline
Models & \#params & Speed & s-acc & t-acc \\
\hline
LEO & 6.9B & 11.3ms & 62.8 & 34.1 \\
\hline
3D-VisTA & 101.1M & 5.2ms & 60.3 & 28.8 \\
 \cellcolor{lightorange}\textbf{3D-VisTA+ GroundFlow}& \cellcolor{lightorange}\textbf{123.1M} & \cellcolor{lightorange}\textbf{5.6ms}&\cellcolor{lightorange}\textbf{64.1} & \cellcolor{lightorange}\textbf{35.1} \\
\hline
PQ3D & 167.4M & 6.8ms & 57.3 & 25.9 \\
 \cellcolor{lightorange}\textbf{PQ3D+ GroundFlow}& \cellcolor{lightorange}\textbf{189.4M} & \cellcolor{lightorange}\textbf{6.9ms}&\cellcolor{lightorange}\textbf{64.8} & \cellcolor{lightorange}\textbf{36.1} \\
\hline
\end{tabular}
\caption{Comparisons of size, speed and performance of different models. \#params indicates the number of parameters each model has and speed shows the inference time per step.}
\label{tab:compute}
\end{table}

\section{Data efficiency}
\begin{figure}[h]
    \centering
    \includegraphics[width=0.8\columnwidth]{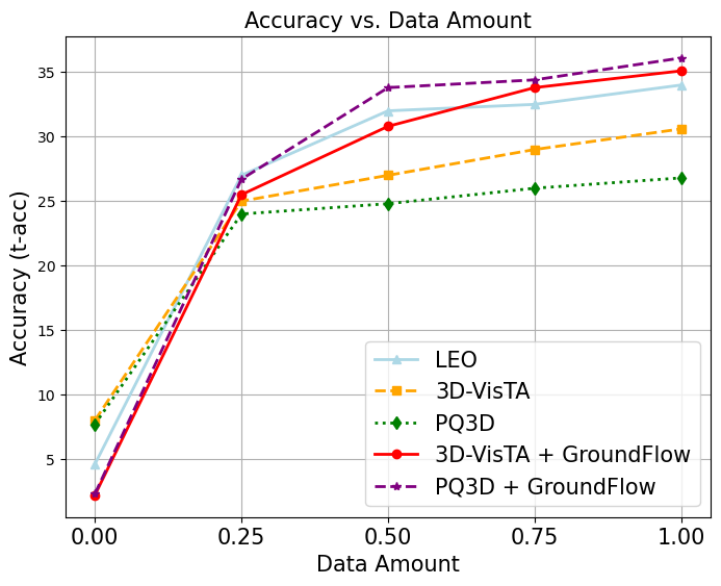}
      \caption{Comparisons of data amount versus task accuracy (t-acc).}
  \label{fig:data_eff}
\end{figure}

Figure \ref{fig:data_eff} shows that increasing the amount of data improves the performance of all methods. Notably, 3DVG methods integrated with GroundFlow demonstrate superior data efficiency. They achieve performance comparable to 3DVG baselines using only 50\% of the data and surpass the 3DLLM LEO model with less than 75\% of the data. This advantage likely stems from GroundFlow's specialized design for sequential grounding task, which enables the model to efficiently learn from historical information in context and generalize to unseen step instructions.

\section{Failure Cases}
\begin{figure}[h]
    \centering
    \includegraphics[width=0.85\columnwidth]{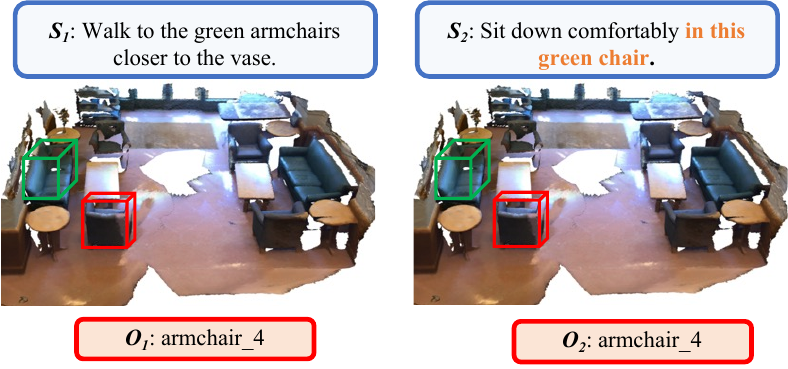}
      \caption{Example of the failure case.}
  \label{fig:failure}
\end{figure}

We found that most failures of GroundFlow occur when an incorrect prediction is made in previous steps. As shown in Figure \ref{fig:failure}, since GroundFlow is built on 3DVG methods, any incorrect information generated by 3DVG methods gets propagated to future steps, leading to wrong predictions.

\section{More Qualitative Comparisons}
\begin{figure*}[h]
    \centering
    \includegraphics[width=0.9\textwidth]{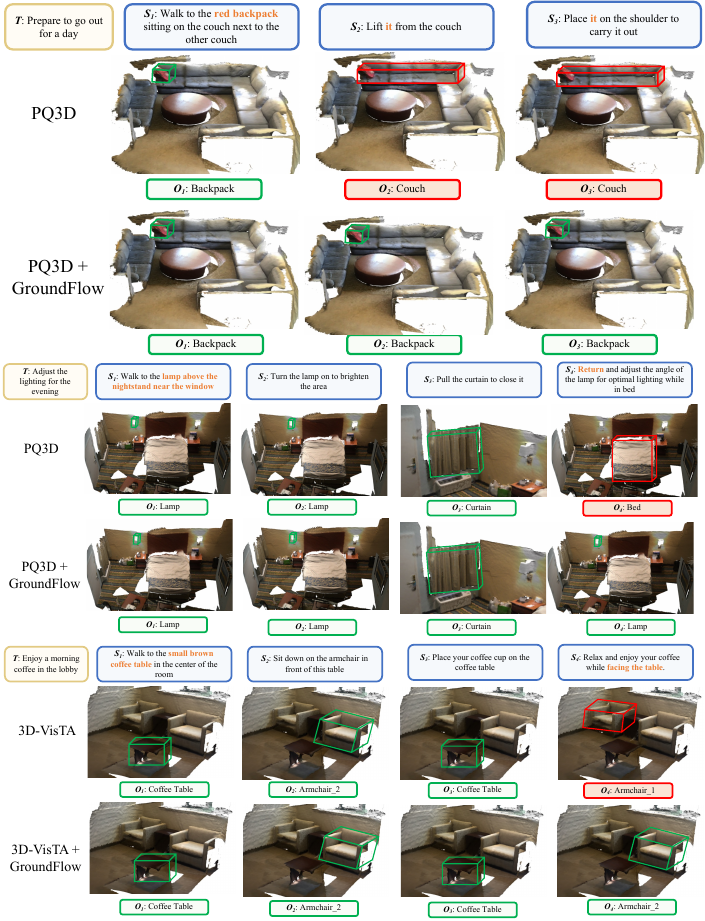}
      \caption{Visualization results from PQ3D and 3D-VisTA without and with GroundFlow. $T$ represents the task description, $S_t$ and $O_t$ denote the step instruction and corresponding referred target object in step $t$. \textcolor{red}{Red} are wrong predictions and \textcolor{green}{green} are correct predictions.}
  \label{fig:supp_visual}
\end{figure*}

We include more qualitative comparisons in Figure \ref{fig:supp_visual}. It is shown that GroundFlow enables the 3DVG baseline models to effectively capture contextual past information to make current step predictions.
\end{document}